\documentclass{amia}
\usepackage{graphicx}
\usepackage[labelfont=bf]{caption}
\usepackage[superscript,nomove]{cite}
\usepackage{color}

\begin{document}

% Or, simply replace dp with t1d
\title{Modeling Disease Progression Trajectories\\ from Longitudinal Observational Data}

% \author{Bum Chul Kwon, Ph.D.$^{1}$, Kenney Ng, Ph.D.$^{1}$, Vibha Anand, Ph.D.$^{1}$}
\author{Bum Chul Kwon$^{1}$, Peter Achenbach$^{2}$, Jessica L. Dunne$^{3}$, William Hagopian$^{4}$,\\ Markus Lundgren$^{5}$, Kenney Ng$^{1}$, Riitta Veijola$^{6}$, Brigitte I. Frohnert$^{7}$, Vibha Anand$^{1}$,\\ and the T1DI Study Group}

\institutes{
    $^1$IBM Research, Cambridge, Massachusetts, United States; $^2$Helmholtz Zentrum M\"unchen, Germany; $^3$JDRF, New York, New York, United States; $^4$University of Washington, Seattle, Washington, United States; $^5$Department of Clinical Sciences, Lund University, Malm\"o, Sweden; $^6$University of Oulu, Oulu, Finland; $^7$University of Colorado, Denver, Colorado, United States
}

\maketitle

\noindent{\bf Abstract}

\textit{Analyzing disease progression patterns can provide useful insights into the disease processes of many chronic conditions. These analyses may  help inform recruitment for prevention trials or the development and personalization of treatments for those affected. We learn disease progression patterns using Hidden Markov Models (HMM) and distill them into distinct trajectories using visualization methods. We apply it to the domain of Type 1 Diabetes (T1D) using large longitudinal observational data from the T1DI study group. Our method discovers distinct disease progression trajectories that corroborate with recently published findings. In this paper, we describe the iterative process of developing the model. These methods may also be applied to other chronic conditions that evolve over time.}

\section*{Introduction}

Analyzing disease progression can provide great insights into the underlying pathophysiology of disease processes.
Examining how a disease progresses over time can help us to understand when a patient will progress into a stage requiring treatment and provide tailored care for the patient.
Numerous natural-history studies have been conducted for specific diseases, such as for type 1 diabetes (T1D), Huntington's disease, and chronic obstructive pulmonary disease (COPD) in the past.
These studies provide great opportunities for clinical researchers to investigate disease progression patterns using longitudinal data.

However, despite the availability of well-curated data from these past observational studies and the best of many research motives, many challenges remain to conduct such investigations. For example, to understand how subjects progress from no disease into disease states, one has to take into account multiple socio-demographic as well as clinical measures and their evolution over time that may be associated with the onset of the disease.
Second, there are challenges inherent to collection of longitudinal data, i.e. even if the data are collected for all co-variate(s) of interest, observations may be recorded at discrete times and varying intervals in many similar studies. 
This introduces sampling bias.
Third, the observational records may also contain missing data, i.e. not each co-variate may be recorded at each time point in the data collection process.
Fourth, depending on the chronic condition,  disease progression rates and patterns may vary indicating  heterogeneity of progression across populations which may be attributed to underlying  environmental and/or genetic risks. 
For these reasons, it is not only difficult to find distinct progression patterns from large observational studies of the past but to also summarize these patterns in an appropriate way for downstream consumption~\cite{Sun2019_JAMIA}. For example, researchers and clinicians may want to infer distinct trajectories from progression patterns for further investigations, and needless to say this largely remains an active area of biomedical research~\cite{steck_predictors_2015, endesfelder_novel_2016, cook_disease_2016, liu_efficient_2015, chen_penalized_2017, wang_unsupervised_2014}. In this paper, we address the question of learning disease progression patterns from noisy observational data using data-driven probabilistic methods and interactive visualization approaches. As a proof of concept, we apply these methods to the multi-site data from the ``T1DI'' (Type 1 Diabetes Intelligence) study group. Our team is involved in this effort with an aim to study computational modeling of T1D. 

T1D is a chronic auto-immune condition that affects both children and adults and has lasting consequences of insulin dependence throughout an individual's life. By recent estimates, the disease incidence is doubling every five years, particularly in children below the age of 5 years, and yet there is no cure for it~\cite{gomez2019correlating}. It is now known, that a presymptomatic phase which involves development of islet autoantibodies (biomarkers) followed by a period of dysglycemia, precedes the onset of clinical symptoms when an individual is generally diagnosed. 
A recent scientific statement by the leading research organizations have proposed three clinical stages of the disease~\cite{insel_staging_2015}.
In the first stage, individuals show presence of multiple islet autoantibodies but with normal glycemic control; in the second stage, in the presence of islet autoimmunity, they start exhibiting dysglycemia; and in the third stage the overt clinical symptoms set in when a clinical diagnosis is generally made. While these clinical stages provide a big picture of disease progression, they do not address its heterogeneity , i.e. when an individual is likely to progress, and what may be the next stage and under what circumstances. However, heterogeneity has been well described in past and ongoing research studies~\cite{hagopian_environmental_2011} as well as in clinical practice. 

To address these questions, we model disease progression using Hidden Markov Models (HMMs), a class of state-space models, that can discover underlying latent (disease) states from observational data in a probabilistic way~\cite{Sukkar2012, wang_unsupervised_2014, Sun2019_JAMIA, LiuLiEtAl2015_NIPS,JacksonEtAl2003_JRSS}. We apply this method iteratively to learn an optimal number of disease states based on available (biomarker) data. We supplement our modeling efforts with an interactive visualization method~\cite{kwon2020dpvis} to facilitate discovery of overall patterns (aka disease trajectories) and their salient characteristics. 

Because of growing body of evidence supporting role of biomarkers in progression to onset ~\cite{endesfelder_time-resolved_2019} , the T1D research community is keenly interested in investigating these. These may include the type or number of islet autoantibodies detected in blood serum or their combinations. It is believed that a more detailed map of these patterns may help in further refining disease staging which in turn may support newer prevention trials and personalized therapies.

\section*{Methods}

We first breifly describe the study data followed by modeling and visualization methods. This study was conducted by approval from the IRB of individual study cohorts as well as by the T1DI Study group.

\subsection*{Data: Longitudinal Observational Studies from Five Different Sites}

We use longitudinal data collected from the T1DI Study Group, which combines observational data from five natural history studies, some of which are still ongoing. These studies are  BABYDIAB~\cite{ziegler1999autoantibody}, DAISY~\cite{rewers_newborn_1996}, DEW-IT~\cite{wion_population-wide_2003}, DiPiS~\cite{jonsdottir_childhood_2018}, and DIPP~\cite{nejentsev_population-based_1999}.
In these studies, each site recruited children at genetic or familial risk at birth or close to birth and followed them in periodic visits until their diagnosis or for a period of 15 years whichever came first. This 
combined and harmonized cohort of five studies has over 24,000 subjects, with an average of 12 (sd: 9) visits per subject, and an average interval of 0.8 (sd: 0.94) years between visits. Of these, 2524 (10\%) subjects developed one or more autoantibodies in the follow up period and 697 (3\%) were diagnosed with T1D (at the time of these analyses).

Of those diagnosed, we analyze 688 T1D cases partitioned into two sets. We use visits of 559 T1D cases from 3 studies (DAISY, DIPP, DiPiS) having at least 3 visits in the follow up period as our modeling set (i.e. for learning a disease progression model). We use 129 T1D cases from 2 studies (BABYDIAB and DEWIT) as an independent evaluation set and label their visits at each time point with the model latent states. These are described in detail below.

\subsection*{Modeling: Training Hidden Markov Models}

An HMM can represent progression (of a subject) through disease stages as transitions between `hidden (or latent) states.'
The hidden states when learned from data characterize a set of probability distributions of multiple observed measures. In these models, state transitions (among latent states) are represented using a transition matrix. The transition matrix defines the probability of transitioning between hidden states and if needed can be constrained based on domain knowledge or pragmatic assumptions. 
For example, one can make an assumption that the disease progresses in one direction, i.e. forward only and a subject does not revert back to prior states as time goes by. 
There are several advantages to using a probabilistic disease progression modeling framework.
First, it provides a flexible framework to accommodate different possible progression pathways.
Second, the trained HMM can infer the best state representations from data, which may include missing data.
Third, it allows multi-dimensional representations of disease states, i.e. evolution of multiple covariates.
Fourth, it infers the best possible progression trajectory of an individual subject and collectively can look at trajectories of an individual or a population.
Finally, this method does not require labeled data or ground truth since the primary goal of modeling is to probabilistically discover underlying patterns by way of latent states , their characteristics and transitions.

In many applications, HMMs are trained using periodic time-stamped data in an unsupervised manner. These applications generally use the discrete-time HMM (DT-HMM) which do not allow for irregular time intervals of measurements in the observed data. However, this irregular sampling is often the case in clinical datasets such as the one at hand. Another variation of HMMs, a continuous-time HMM (CT-HMM) can accommodate varying time intervals though with added computational complexity.

In our prior work on Huntington's Disease (HD)~\cite{Sun2019_JAMIA}, we have successfully used CT-HMMs to learn the underlying disease states of HD, a pure genetic condition, purely based on observed clinical assessments. In this paper, we leverage the previous work and apply it to the domain of T1D to understand progression from presymptomatic phases to the onset of disease.  However, we seek some specific changes for our use case. First, we extended the observed variable space from continuous measures (of clinical assessments) to include categorical measures of biomarkers. This was important in this application because the presence or absence of a biomarker (islet autoantibody) detected in blood-serum are indicative of progression and in fact has been proposed as stage 1 of the condition in a recent joint scientific statement~\cite{insel_staging_2015}. Additionally, it is now believed that the type of autoantibody detected in blood serum first may also define the progression pattern to onset of the disease~\cite{krischer20156}. Therefore, it made sense to use categorical measures of these biomarkers (i.e. positive or negative) in our models. This allowed us to leverage the standardized cutoffs of autoantibody titers without having to first harmonize these lab values across these past studies of many years. 
Secondly, we apply a principled approach to model selection, i.e. for determining the number of latent states that may optimally define progression. It takes into account both the information criterion and log-likelihood (LL) scores.
In generative models, like HMM, LL is often used to measure the goodness of fit for model selection, i.e. the probability of the observed data given the model.
However, LL does not penalize for model complexity, i.e. the number of parameters to learn or the number of observations required.
In this work, besides using LL to guide the model selection, we also evaluate the Bayesian Information Criteria (BIC) score. BIC penalizes for the number of model parameters and the number of observations~\cite{claeskens2008model}. 

In summary, for training HMMs in the context of disease progression, clinical researchers have an ardent task: i) to choose a set of observed variables, ii) to determine the number of latent states (\textit{K}) for the model, iii) to set constraints on transition probabilities (between latent states) in a pragmatic way, iv) to select a model based on criteria scores and v) to qualitatively evaluate the model so that it may meaningfully define progression for clinical consumption.
% While tasks i) to iv) relate to the modeling work, task v) relates to our visualization work that is currently under review. 
We attempt to address these challenges in this work by taking a unique approach of combining modeling and visualization in an iterative way which we describe next.

\textbf{Model Setup: Observations, Number of States, Transition Constraints, and Model Outputs}

First, we needed to choose a set of observed variables to discover the hidden states from. In our use case, we use previously recognized islet autoantibody biomarkers: glutamic acid decarboxylase autoantibodies (GADA), insulinoma-2-associated autoantibodies (IA-2A), insulin autoantibodies (IAA) that are known for their association to T1D disease status.

Second, we needed to choose the number of hidden states (\emph{K}) for the HMM model. This model selection process is done by building models with different \emph{K}, quantifying the model characteristics, comparing the models, and selecting an appropriate \emph{K}. Bootstrap resampling was used to assess the model fit as measured by LL. To supplement the model selection process, a variety of other methods were applied including BIC, visualization and model constraints. For example, for each model, the latent states are characterized by probabilities (distribution) of observed variables and these were explored in the context of model explainability using visualization methods~\cite{kwon2020dpvis}.
In general, as the number of latent states increase, more transition pathways are possible between them.
While these pathways may expose heterogeneity in disease progression, they can also increase model complexity and limit model interpretability. Typically, models with high \emph{K} can have lower error rates (e.g., higher log-likelihood scores, lower BIC scores) but are more complex and harder to interpret.

\begin{figure}[bth]
\centering
\includegraphics[width=\textwidth]{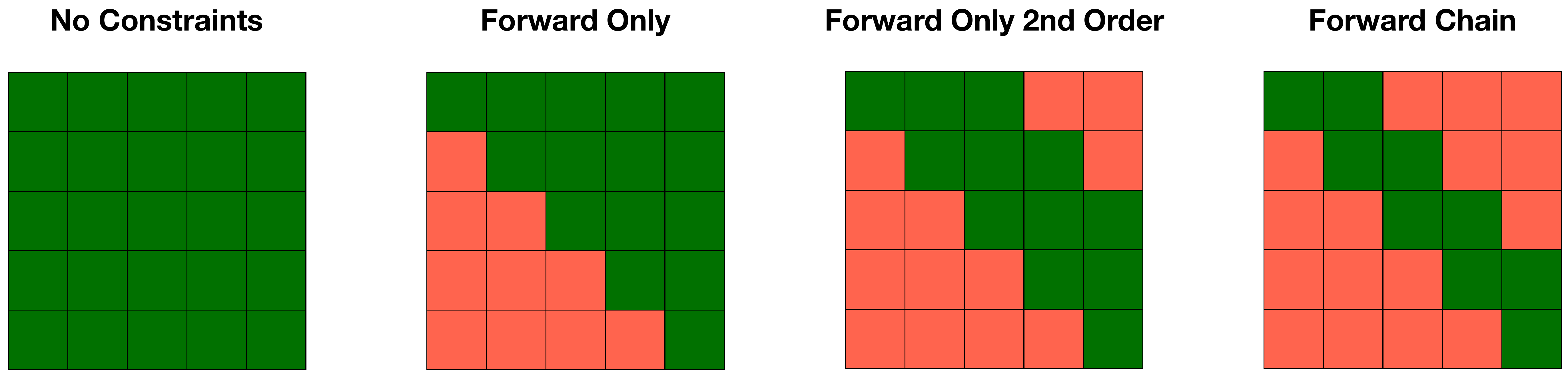}
\caption{The types of constraints that can be set on transition probabilities of Hidden Markov Models (HMMs). Each constraint is represented as a $K$$\times$$K$ matrix, where $K$ is the number of states. The four examples illustrate constraints as green (possible route between states) and red (restricted route between states). }
\label{fig:hmmconstraints}
\end{figure}

Another way to increase model interpretability is by imposing state transition constraints.
However, this can depend on the nature of the disease. 
Figure~\ref{fig:hmmconstraints} illustrates various kinds of constraints that can be set on the transition probabilities.
No constraints can be used to allow transitions between all states.
On the other hand, forward only constraints can be set so that transitions in the reverse order are not allowed.
This is especially helpful to capture chronic diseases that exhibit worsening symptoms as time goes by. 
In addition to directionality, constraints can also be set on the transition steps.
For instance, to represent the step-by-step progression nature, we can set the model to be a forward-chain model, which means a patient can only either stay at state $i$ or jump to state $i+1$.

Once we have a trained model, several model outputs can be assessed to interrogate the model and draw useful insights. First, as a result of the training process, a (latent) state is assigned to every visit encounter in the modeling set. Essentially this assignment results in ``labeling'' of longitudinal visits as temporal sequences of HMM states.
In addition to state sequences, we also derive posterior probabilities (of state assignment) from the model (just considering forward pass) for every visit. The Viterbi method (that relies on forward-backward pass) employed in our models, picks the state sequence with the highest probability~\cite{forney1973viterbi}. In case of discrepancies between the two methods, the output shows uncertainties of the assignment process. Potentially these can be visualized to derive useful insights into both the modeling and the resulting progression.
Other useful insights can be drawn from the finite number of states (of the model). These can potentially be used to group subjects with similar state sequences. We can then compare the heterogeneity with respect to progression rates using interactive visualizations. Other useful outputs from these models are a state transition matrix that defines the rate of transition from state $i$ to state $j$. This rate can be used to compute average dwell times (sojourn times) in each state and the probability of staying in or transitioning out from a state given a time value of interest (e.g. in 2 years).

\subsection*{Model Evaluation: Assessing Hidden Markov Models}

\begin{figure}[t!]
\centering
\includegraphics[width=\textwidth]{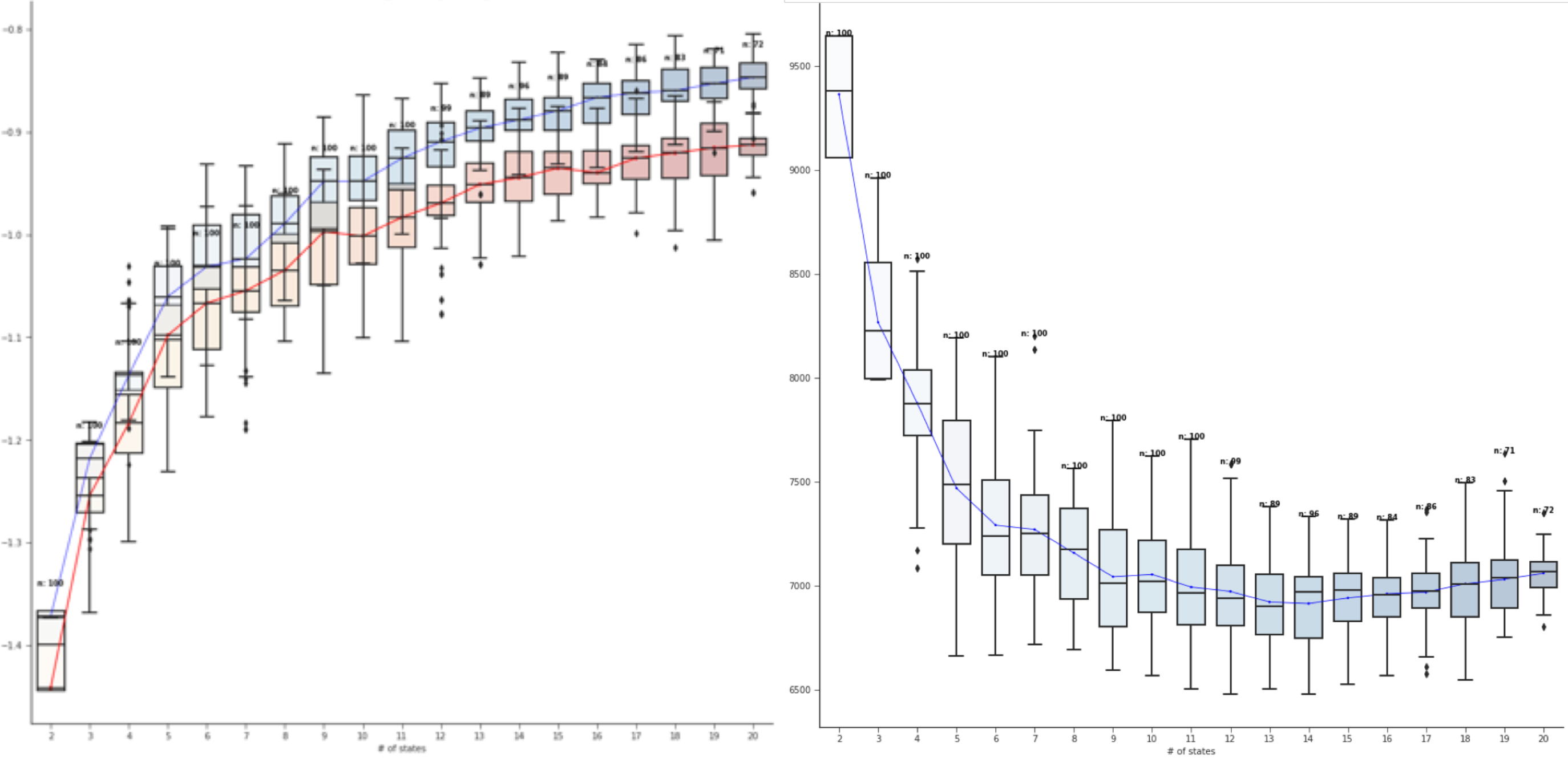}
\caption{Model performance (vertical axis) as a function of the number of latent states $K$ (horizontal axis). The left view shows the model performance in terms of predictive likelihood score (LL) on the validation set; the right view shows the model performance in terms of Bayesian Information Criterion (BIC) on validation set.}
\label{fig:hmmevaluation}
\end{figure}

To find the optimal model, we needed to run experiments with different combinations of parameters for HMMs (number of states, transition constraints) and impose visualization. For the first, we randomly split data into training and validation sets using the 7$:$3 ratio. We created V such experimental sets. Using each experimental setup, in our experiments, we varied the number of latent states from $K=2$ to $K=20$ states, i.e. we learned a model for each latent state by random initializations. We did not consider any numbers beyond 20 (for latent states) because of complexity in terms of the possible number of transitions and interpretability.
We chose the forward chain model, where subjects can either stay at the current state ($i$) or proceed to the next state ($i+1$) while prohibiting them from jumping forward or backward into other states. However each subject could start or end their progression in any state. 
Thus, each experimental setup was randomly initialized 10 times and in each initialization a new HMM was trained using the iterative expectation–maximization (EM) algorithm~\cite{wang_unsupervised_2014}. At each iteration, a training (set) LL was computed to assess model convergence. A LL score on the validation set was similarly computed at the end of convergence and is used as a performance metric (of predictive LL) for model evaluation and selection as described below. 
We plot both the training LL and the predictive LL as a function of the latent states used in these experiments.
The total number of experiments (\emph{N}) in the above setup amounts to the product of the number of data splits (\emph{V}), the number of candidate latent states (\emph{K}), the number of experiments (\emph{M}), and the types of constraints (\emph{C}). In total, we ran the experiments for training 1900 model instances (V:10, K:2 to 20 states, i.e. total of 19 states, M:10, and C:1) which are plotted in Figure~\ref{fig:hmmevaluation}.

\subsection*{Visualization Method: Summarize States and Transition Patterns}

To summarize the trained model, we apply DPVis~\cite{kwon2020dpvis}, an interactive visualization tool developed for users to explore disease progression patterns.
DPVis is a visual analytics system that consists of multiple, coordinated views, where each view provides summary and details of the HMM states and the state transition patterns.
The visualization has three main goals.
First, it helps users examine the summary of states by inspecting the probabilities of output variables per state.
In this process, users can also evaluate the probabilities of other variables (per state), which though were unused in the modeling process.
Second, it allows users to summarize the discovered longitudinal clusters with respect to state transitions, i.e., \emph{trajectories}.
Users can find the volume of individuals and compare heterogeneity in terms of progression rates by viewing these trajectories.
Third, it allows users to further group individuals based on state transition patterns.
Users can form and test hypotheses by interactively forming subgroups and examining their characteristics.
DPVis has many additional functionalities and views designed to help facilitate investigation of disease progression patterns, but a comprehensive and detailed description of them is out of scope here and described elsewhere~\cite{kwon2020dpvis}.
In this paper, we apply DPVis to a specific use case to help generate summaries of HMM states and state transition patterns over time, as shown in Figures~\ref{fig:hmmoutputvariables} and~\ref{fig:statetransitions}.

\section*{Results}
Model performance as a function of the number of latent states is shown in Figure~\ref{fig:hmmevaluation}.
The left plot shows the LL of the training dataset (at the last EM iteration) (blue) and validation set (red) (after model convergence) over different numbers of latent states $K$ (horizontal axis).
For each $K$, a box plot shows the summary of 100 experiments (10 random training/validation splits used with 10 random model initializations).
We have connected the median results with a line to show the trend.
Figure~\ref{fig:hmmevaluation} shows that the LL score in general increases as the number of states increase, which is expected.
However, the magnitude of increase in LL starts decreasing around 9--11 states, which indicates that the models with more number of states than 9--11 states may not add value for the cost (complexity).
To validate our observations, we computed BIC on the validation set, which is shown in the right plot of Figure~\ref{fig:hmmevaluation}.
The BIC score decreases as the number of states increases and is consistent with the trend observed with LL. 
We observe that the BIC measure starts flattening out as the number of states increase beyond 11--13 states. 
By combining the evaluation results from two independent measurements, LL and BIC, we chose the 11 state model, which has a good trade-off between good model fit and low model complexity.

\begin{figure}[t!]
\centering
\includegraphics[width=\textwidth]{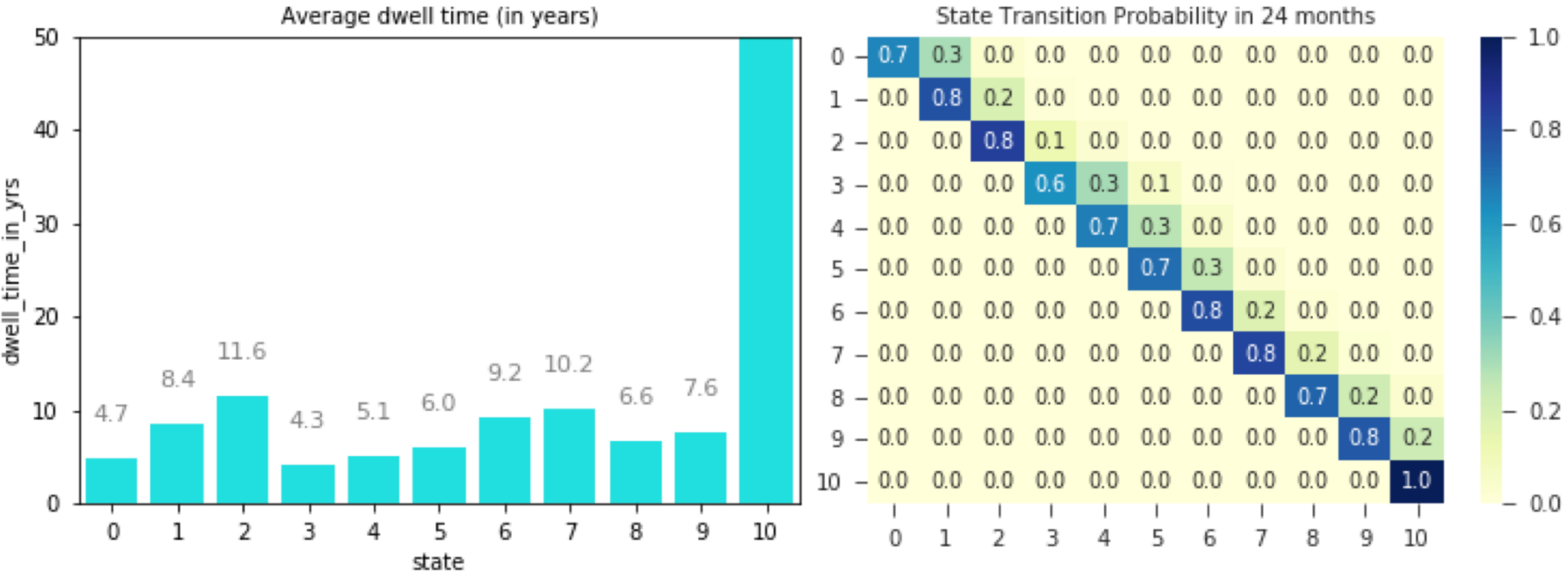}
\caption{The trained model provides various outputs. The left plot shows the average dwell time per state. The right plot shows the transition probabilities among states in 2 years.}
\label{fig:hmmdwelltransition}
\end{figure}

Once a model instance has been trained, its outputs (e.g. dwell time, transition matrix) can be interpreted both quantitatively and qualitatively. 
Figure~\ref{fig:hmmdwelltransition} shows two different outputs of a trained model: dwell time (left) and transition probabilities (right).
Average dwell time indicates the average time subjects stay at each state. 
The bar chart shows some patterns with state sub-sequences. 
We observe that average dwell time increases as the state index increases until some point, then decrease to small numbers, and then increases back until some other points.
In total, we observe three different segments: i) 0--2, ii) 3--7, iii) 8--10.
Within each segment, we observe that average dwell time increases as the state index increases. 
We also note that state 10 is the ``sink'' state by the construction of the model (i.e., it is the last state).
The transition probabilities show patterns for how subjects transition from one state to another in a given time frame. 
Figure~\ref{fig:hmmdwelltransition} (right) shows the state transition probabilities (after 24 months) as a matrix.
As expected, after 24 months, most subjects stay at their current state and rarely move to next states as shown by the large numbers in the diagonal cells.
% Average dwell time shows some patterns of increasing dwell times.

\begin{figure}[b!]
\centering
\includegraphics[width=\textwidth]{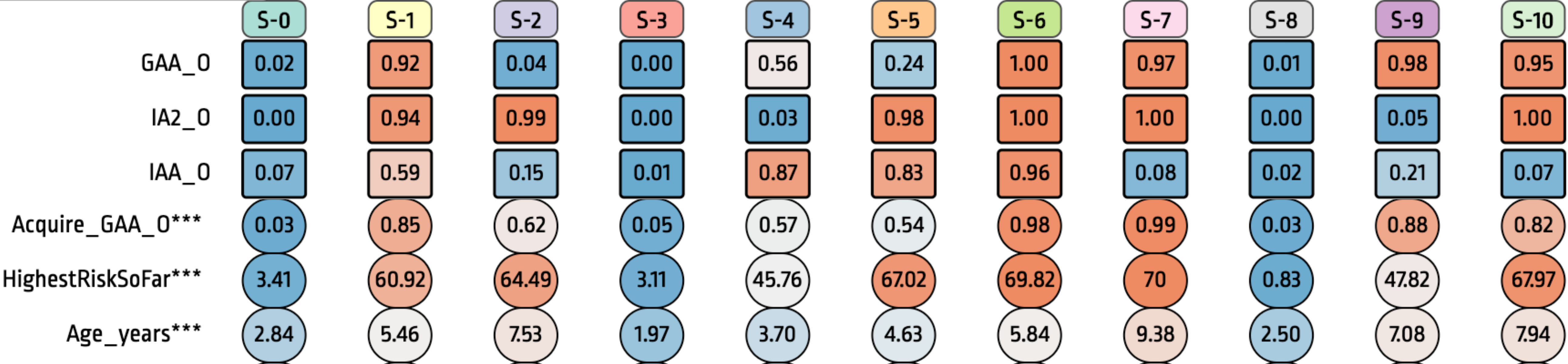}
\caption{The Summary of HMM Output Variables. Each state (column) shows probabilities of onsets of three output variables (rows): GADA, IA-2A, and IAA, which were used for modeling. In addition, it shows probabilities of other variables that are unused for the modeling. Acquire$\_$GAA shows the probability of acquiring GADA in the corresponding state, HighestRiskSoFar is the mean value of the derived measure which shows the maximum risk score each subject has before arriving in the corresponding state, and Age$\_$years shows the mean ages of subjects in the corresponding state.}
\label{fig:hmmoutputvariables}
\end{figure}

\begin{figure}[t!]
\centering
\includegraphics[width=\textwidth, trim={.0cm 15.0cm 0.0cm 0cm},clip]{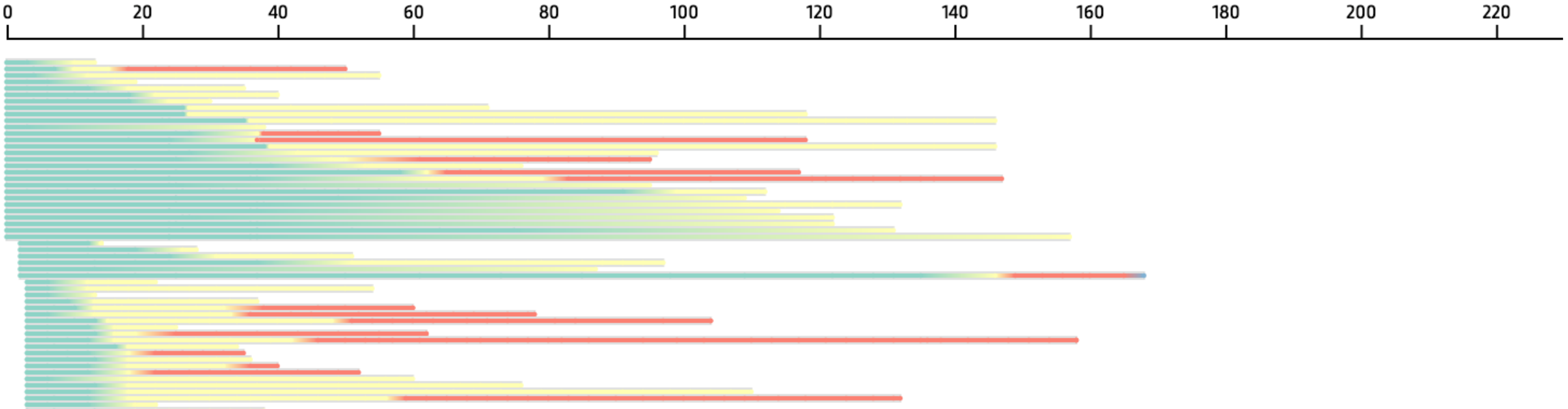}
\caption{State transition patterns are shown as temporal summary. Each row represents a subject's timeline (horizontal axis, in months), where each color band indicates the subject's dwell time in the corresponding state (state 0: green; state 1: yellow, state 2: red).}
\label{fig:statetransitions}
\end{figure}

There are several outputs from the HMM that can be visualized to draw insights. One of them is the state sequence of the trained model applied to visit observations from individual subjects in the training, validation or independent datasets.
We apply DPVis for exploring these outputs~\cite{kwon2020dpvis} to gain an understanding of progression patterns captured by the trained model.
Figure~\ref{fig:hmmoutputvariables} shows a summary of the selected 11-state model when applied to the training, validation and independent datasets.  
Consistent with what we observed looking only at the average dwell times, we found three segments for state transition probabilities. In terms of observed variables used in the model (IAA, GADA, and IA2A) and their distributions, we can confirm that each segment, which we are calling a ``trajectory,'' follows different state transition patterns.
The first trajectory includes three states: 0, 1, and 2. 
State 0 has low probabilities of all three islet autoantibodies (state 0), which are followed by high probabilities of all three (state 1), and then high probability of IA-2A (state 2).
The second trajectory includes five states: 3, 4, 5, 6, and 7.
State 3 has low probabilities of all three islet autoantibodies, followed by states which add islet autoantibodies IAA, IA-2A, and GADA one by one (states 4,5,6 respectively), and then lose IAA at the end (state 7).
Finally, the third trajectory starts with low probabilities of all three islet autoantibodies (state 8), followed by high probability of GADA (state 9) and high probabilities of GADA and IA-2A (state 10).

We can further visualize the heterogeneity of the three trajectories using DPVis.
Figure~\ref{fig:statetransitions} shows an overview of state transition patterns as a function of the age of subjects (horizontal axis).
Each row represents a subject's timeline and each color band represents the subject's dwell time in the corresponding state.
We filtered the view to focus on the first trajectory pattern (state 0 (green), 1 (yellow), and 2 (red)).
Although all subjects follow the same state transition pattern (0--1--2), the plot reveals that each subject makes state transitions at different ages.
For example, dwell times for state 0 (green) are very different between subjects.
The first subject includes a very short state 0 dwell time (less than a year) and there are subjects who dwell in state 0 for more than 10 years.
The subsequent states, 2 and 3, also show heterogeneity in terms of entering ages and dwell times. We can draw similar insights from the other two trajectories.

\section*{Discussion \& Conclusion}

We present an analytic approach that combines the power of modeling with HMMs and interactive visualization techniques for application in longitudinal observational datasets. We show that this approach can draw meaningful insights for understanding disease progression even when data had been collected at irregular intervals and is categorical by nature. Furthermore, we describe various aspects of modeling that one may need to consider with HMMs and how interactive visualization can supplement in this effort. In particular, we applied the above methods to the domain of T1D datasets using islet autoantibodies as biomarkers of disease progression.

As a result of training multiple models with different parameters, and using a principled approach for model selection, we found that an 11-state model showed good model fit, low error rates, and low model complexity, at least for these datasets.  In addition, the model captured patterns (of islet autoantibody progression) that are similar to some recent reports~\cite{steck_predictors_2015, endesfelder_novel_2016, endesfelder_time-resolved_2019}. This is important as it speaks to the usefulness of our approach, especially when these reports are from large multi-site NIH funded ongoing studies (of T1D) that are collecting data at more frequent and regular intervals, also to understand the heterogeneity of disease among many other aims. Needless to say that our approach may be highly desirable for other similar chronic conditions that have been studied in past natural history studies.

Additionally, we have shown the value of interactive visualization to summarize disease progression patterns. In the following qualitative evaluation using interactive visualizations, we have shown that the HMM model is able to provide a population-level summary of state transition patterns. These are the three trajectories and details in terms of dwell times and transition ages as discussed above. These insights may have practical applications that are yet to be understood. Therefore, as next steps, we are evaluating many of the model outputs (state characteristics) in more detail. At the same time, we are also working on advanced modeling (using a combination of categorical and numerical  variables) as well as constraining the model in other useful ways and on enhancing the visualization approach with many features to better support a clinically focused set of users. We plan to disseminate this work in future.

\section*{Acknowledgment}

We wish to thank the T1DI Study Group for their help in this work and study participants of the other two studies.
The T1DI Study Group consists of following members: 1) JDRF--Frank Martin, Jessica Dunne, Olivia Lou; 2) IBM--Vibha Anand, Mohamed Ghalwash, Eileen Koski, Bum Chul Kwon, Ying Li, Zhiguo Li, Bin Liu, Ashwani Malhotra, Kenney Ng; 3) DiPiS--Helena Elding Larsson, Josefine J\"onsson, \AA{}ke Lernmark, Markus Lundgren, Marlena Maziarz, Lampros Spiliopoulos; 4) BABYDIAB--Peter Achenbach, Christiane Winkler, Anette Ziegler; 5) DIPP--Heikki Hy\"oty, Jorma Ilonen, Mikael Knip, Jorma Toppari, Riitta Veijola; 6) DEW-IT--Bill Hagopian, Michael Killian, Darius Schneider; 7) DAISY--Brigitte Frohnert, Jill Norris, Marian Rewers, Andrea Steck, Kathleen Waugh, Liping Yu.
This work was supported in part by JDRF (1-IND-2019-717-I-X, 1-SRA-2019-722-I-X, 1-SRA-2019-723-I-X, 1-SRA-2019-719-I-X, 1-SRA-2019-721-I-X, 1-SRA-2019-720-I-X).

\makeatletter
\renewcommand{\@biblabel}[1]{\hfill #1.}
\makeatother

\bibliographystyle{unsrt}
\bibliography{main}

\end{document}